%% file: main.tex
\documentclass[10pt,twocolumn,letterpaper]{article}

\usepackage{cvpr}              
\usepackage{booktabs}
\usepackage{array}
\usepackage{pifont}
\usepackage{booktabs}
\usepackage{enumitem}
\usepackage{algpseudocode}
\usepackage{subcaption}
\usepackage{graphicx}
\usepackage{amssymb}  
\usepackage{amsmath}
\usepackage{longtable}
\newcommand{\xmark}{\ding{55}} 
\usepackage{enumitem} 
\usepackage{graphicx}
\usepackage{multirow}
\usepackage{amsmath}
\usepackage{subcaption}
\usepackage[ruled,vlined]{algorithm2e}

\newlist{inline}{enumerate*}{1}
\setlist[inline]{before=\unskip{: }, itemjoin={{; }}, itemjoin*={{; and }}, label={(\roman*)}}

\newcommand{\dmerge}{d_{\text{merge}}}
\newcommand{\ccenter}{c_{\text{center}}}
\input{preamble}
\definecolor{cvprblue}{rgb}{0.21,0.49,0.74}
\usepackage[pagebackref,breaklinks,colorlinks,allcolors=cvprblue]{hyperref}

\def\paperID{*****} 
\def\confName{EarthVision}
\def\confYear{2026}

\title{CAFOSat: A Strongly Annotated Dataset for
Infrastructure-Aware CAFO Mapping Using
High-Resolution Imagery}

\author{Oishee Bintey Hoque\thanks{Equal contribution.}$^{*,1}$,
Nibir Chandra Mandal$^{*,1}$,
Mandy L Wilson$^{2}$,
Samarth Swarup$^{2}$,\\
Madhav Marathe$^{1,2}$,
Abhijin Adiga$^{2}$ \\
$^{1}$University of Virginia \quad
$^{2}$Biocomplexity Institute, University of Virginia \\
{{\tt\small oishee@virginia.edu},~{\tt\small wyr6fx@virginia.edu}}
}

\begin{document}
\maketitle
\input{sec/0_abstract}    
\input{sec/1_intro}

\input{sec/2_relatedwork}
\input{sec/3_data}
\input{sec/4_experiments}
\input{sec/5_Discussion}
{
    \small
    \bibliographystyle{IEEEtran}
    \bibliography{main}
}
\input{sec/X_suppl}
\end{document}

%% file: sec/0_abstract.tex
\begin{abstract}
Concentrated Animal Feeding Operations (CAFOs) pose substantial environmental and public-health risks, while also serving as critical nodes in disease surveillance and climate resilience planning. Yet scalable mapping of CAFOs remains difficult due to heterogeneous data sources, noisy point locations, inconsistent annotations, and systematic under-reporting. We introduce \textit{CAFOSat}, a strongly annotated, infrastructure-aware dataset for CAFO mapping across the United States. CAFOSat integrates high-resolution NAIP imagery with multi-source, multi-state CAFO location inventories and converts weak publicly available CAFO geo-locations into refined point annotations through a human-in-the-loop pipeline that combines a learned “AI annotator” with GradCAM-based localization and geometric clustering. We further curate hard negative samples using land-cover–guided sampling with spatial exclusion buffers, and provide infrastructure-level labels (e.g., barns, manure ponds, grazing presence) via manual verification. The resulting dataset contains over $45,000$ image patches spanning 20 states and four major CAFO. Benchmarking across modern CNN, transformer, and vision-language models demonstrates that refined annotations and curated negatives materially improve performance and generalization. Finally, we introduce a synthetic augmentation pipeline that generates realistic CAFO-like variations to diversify training data, yielding additional gains in robustness under distribution shift.
Dataset: \url{https://huggingface.co/datasets/oishee3003/CAFOSat}, Code: \url{https://github.com/oishee-hoque/CAFOSat}.
\end{abstract}

%% file: sec/1_intro.tex
\section{Introduction}
\begin{figure*}[ht]
    \centering
    \includegraphics[width=.9\linewidth]{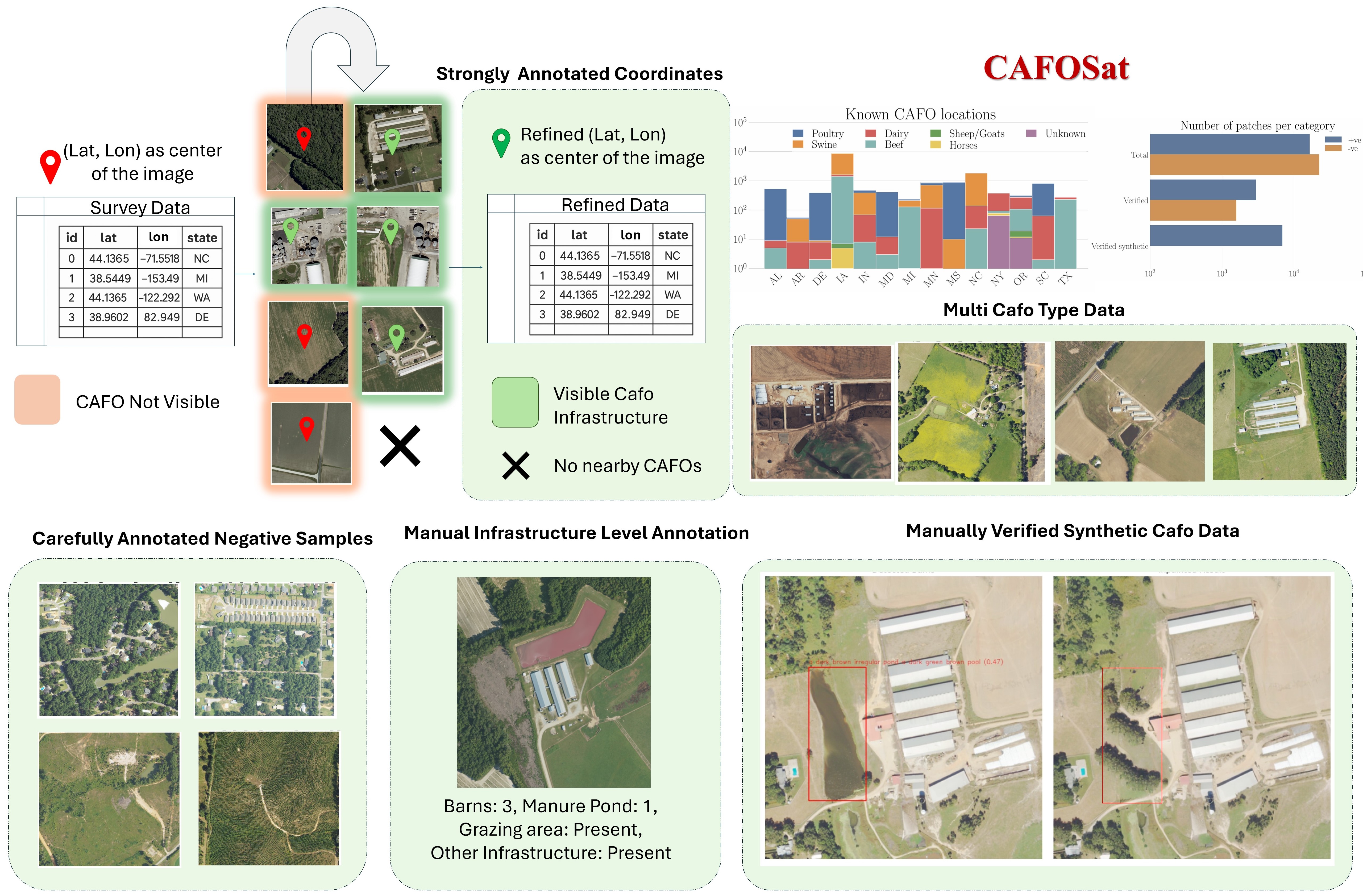}
\caption{Overview of CAFOSat. (\textit{Top left}) Raw survey coordinates (red pins) 
serve as initial patch centers; patches with no visible CAFO infrastructure are 
excluded ($\times$). (\textit{Top center}) The refinement pipeline produces strongly 
annotated coordinates (green pins) centered on confirmed CAFO infrastructure. 
(\textit{Top right}) State-wise CAFO location counts by livestock type and patch 
counts across Total, Verified, and Verified Synthetic subsets; aerial examples 
illustrate visual diversity across CAFO types. (\textit{Bottom left}) Curated 
hard-negative samples from non-CAFO land cover. (\textit{Bottom center}) 
Infrastructure-level annotation recording barn count, manure pond count, grazing 
area, and other structures. (\textit{Bottom right}) Manually verified synthetic 
patches with bounding boxes highlighting inpainted infrastructure regions.}
    \label{fig:dataset_example}
\end{figure*}

Concentrated Animal Feeding Operations (CAFOs) are large-scale industrial 
facilities where substantial numbers of animals are confined in relatively small 
areas~\cite{moses2017industrial,9832662}. The thresholds for classifying an 
operation as a CAFO vary by livestock type—for example, hundreds of cattle or 
millions of chickens. With the continued rise in CAFOs across the United States (US), concerns
have grown regarding their impact on the environment and public health,
particularly within the One Health framework. CAFOs are susceptible to a wide
range of infectious diseases that can spill over to humans, and they generate
vast quantities of waste, contributing to nutrient runoff, degraded water 
quality, and air pollution through greenhouse gas 
emissions~\cite{moses2017industrial}. The ongoing Highly Pathogenic Avian Influenza (HPAI) epidemic exemplifies the 
risks associated with high-density livestock operations~\cite{prosser2024using,humphreys2020waterfowl,adiga2024high,nguyen2025emergence}. 
Beyond simply identifying CAFO locations, mapping additional attributes such as
facility size, livestock type, and structural components—e.g., barns, manure 
ponds, and storage areas—is critical for accurate risk assessment and epidemiological modeling. CAFOs account for over half of US livestock 
production, yet their locations and prevalence remain difficult to track due to regulatory gaps and limited transparency~\citep{GurianSherman2008,Hribar2010, HandanNader2021}. 
Over the years, there have been several efforts from various entities towards identifying 
CAFO locations~(see Handan-Nader~et~al.~\cite{handan2019deep} for a comprehensive review).

Remote sensing-based methods are popular for mapping and monitoring various agricultural 
assets at different spatial scales. 
In the context of livestock, mapping grasslands, livestock operations, and monitoring animals are some
examples of problems that have been examined using remote-sensing methods.
In recent years, several studies have explored large-scale mapping of CAFOs using remote sensing 
and machine learning~(ML). Deep learning–based methods, in particular, have shown strong 
potential for accurately identifying CAFO locations over broad spatial 
extents~\cite{handan2019deep,9832662,zhu2022meter,CHUGG2021102463}. However, these efforts are 
often limited in geographic scope, the range of livestock types covered, or both. To support 
national-scale mapping and to enable estimation of key facility-level attributes, a comprehensive, 
ML-ready dataset is needed. This requires 
\begin{inline}
    \item aggregating all available data on CAFO facility attributes into a high-fidelity database
    \item benchmarking established methods, especially in out-of-distribution settings, 
    to reveal key data gaps and modeling challenges.
\end{inline}



\textbf{Challenges.} 
Despite the need for regulation of CAFOs, there is a lack of
comprehensive, well-aligned datasets. First, the data has to be assembled
from diverse sources~(see Table~\ref{tab:cafo-sources}) and the coverage is small,
spanning a small number of states, resulting in limited geographic
diversity and poor generalization to new regions with different facility 
layouts or land use patterns. The type and amount of
information differs widely across these sources.
Second, available annotations are often inaccurate or imprecise, typically consisting of point coordinates from regulatory filings that may refer to parcel centroids or access roads rather than the actual facility footprints.
Third, under-reporting is common, especially for unpermitted or 
small-scale operations, introducing systematic false negatives that
degrade training quality.
Fourth, the visual variability of CAFOs—across animal type, geography, and season—is rarely captured, limiting the representativeness of the training data. Finally, manually refining and validating these locations to align them with imagery is labor-intensive and difficult to scale, posing a bottleneck to creating high-quality training datasets.

\textbf{Contributions.} We address the limitations of current CAFO datasets through a comprehensive framework that combines weak supervision, data refinement, and human-in-the-loop validation to create a \textit{\textbf{high-quality, large-scale dataset for CAFO detection}}. Our approach significantly reduces manual annotation effort while improving label quality. The resulting dataset is diverse, scalable, and supports benchmarking across multiple CAFO types and geographies. Our key contributions are:

\begin{itemize}
     \item \textbf{Benchmark dataset:} We introduce a refined, strongly annotated CAFO dataset spanning 20 U.S. states, containing more than 45,000 NAIP image patches (39,257 base + 6,454 synthetically augmented) at 833$\times$833 pixels with 0.6m spatial resolution, and covering four major CAFO types along with well curated negative samples. The dataset is compiled from several distinct sources, including federal, state, and academic records, ensuring diverse geographic and operational representation. In addition to facility-level classification, we provide infrastructure-level annotations (e.g., barns, manure ponds) that are manually labeled and verified. It also includes a manually validated synthetic subset, enabling robust evaluation of both real and augmented data. The dataset features both positive samples and a curated set of hard negatives to support comprehensive model training and testing (see Fig.~\ref{fig:dataset_example}).
     \item \textbf{Scalable Annotation Pipeline:} To aid in this process, we develop a generic human-in-the-loop pipeline to refine weakly annotated geolocation data into strongly annotated labels using a scalable, easily extensible framework that significantly reduces manual annotation effort.
     \item \textbf{Comprehensive Evaluation of Annotation Quality:} We demonstrate the utility of our dataset through a series of experiments that highlight the impact of annotation quality and dataset composition on model performance. We also evaluate cross-dataset generalization, training on CAFOSat and testing on external datasets. 
     
     


\item \textbf{Reproducibility and Accessibility:} To facilitate adoption and reproducibility, we release all code, data processing scripts, annotation tools, and data loaders as part of a complete pipeline. The accompanying Jupyter notebooks provide ready-to-use examples for training, evaluation, and visualization, lowering the barrier to entry for researchers and practitioners.
\end{itemize}

\begin{figure*}
    \centering
    \includegraphics[width=\linewidth]{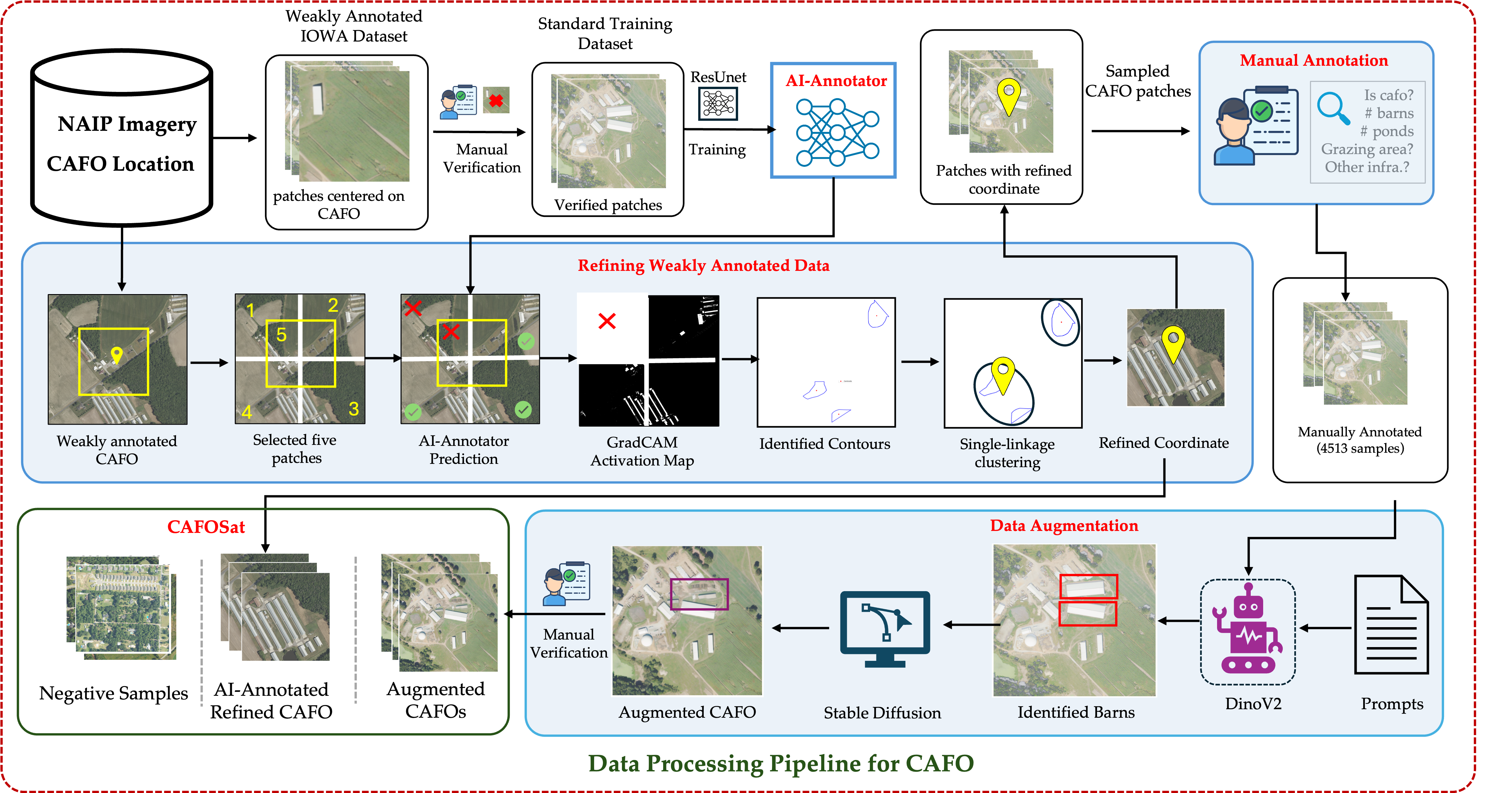}
    \caption{Data processing pipeline for CAFOSat. (\textit{Top}) An AI-Annotator
    trained on verified patches is used to pre-filter CAFO candidates for manual
    annotation. (\textit{Middle}) Weakly annotated locations are refined via five
    overlapping patch extractions, GradCAM activation mapping, contour extraction,
    and single-linkage clustering to produce spatially accurate coordinates;
    4,513 samples are then manually verified with infrastructure-level annotations
    (e.g., barn counts, manure ponds, grazing areas). (\textit{Bottom right})
    A prompt-guided augmentation pipeline uses text prompts and DINOv2 to identify
    infrastructure (e.g., barns), which is subsequently inpainted using Stable
    Diffusion and manually verified. (\textit{Bottom left}) The final CAFOSat
    dataset comprises three components: carefully curated negative samples,
    AI-annotated refined CAFO patches, and manually verified synthetic augmented
    patches, all accompanied by manually annotated infrastructure-level
    information.}
    \label{fig:data_process}
\end{figure*}

%% file: sec/2_relatedwork.tex
\section{Related Work}


\begin{table*}[ht]
\footnotesize
\caption{Comparison of existing ML-ready datasets for CAFO identification.}
\label{tab:cafo-dataset-comparison}
\centering
\renewcommand{\arraystretch}{1.2}
\setlength{\tabcolsep}{11pt}
\begin{tabular}{lllll}
\toprule
\textbf{Dataset} & \textbf{Coverage} & \textbf{CAFO Types} & \textbf{Infrastructure} & \textbf{\# CAFOs} \\
\midrule
METER-ML \cite{zhu2022meter}     & NC, MI, CA, MN   & \xmark                           & \xmark                       & 6,979  \\
RegLab \cite{handan2019deep}     & North Carolina   & Swine/Poultry                    & \xmark                       & 4,984  \\
MEJ-Dataset \cite{ehrenpreis2021nmcafo} & New Mexico & Cattle                          & \xmark                       & 160    \\
\hline
\textbf{CAFOSat (Ours)} & \textbf{20 U.S. states} & Swine/Poultry/Cattle/Dairy/Mixed & \textbf{Barn, Pond, Grazing} & \textbf{18,441} \\
\bottomrule
\end{tabular}
\end{table*}

\paragraph{Remote sensing and CAFO Detection: datasets and methods.} 
Robinson~et.~al.~\cite{9832662} demonstrate that deep learning is a
promising approach to map poultry operations at scale. They do this by training convolutional neural network (CNN) 
models on USDA NAIP 1m aerial imagery, using data from portions of four
states for training and validation. Hendan~et~al.~\cite{handan2019deep} apply
standard CNNs in their research, but
coverage is restricted to swine and poultry operations in North Carolina.
Similarly, the Mapping for Environmental Justice
(MEJ)~\cite{ehrenpreis2021nmcafo} dataset is confined to cattle CAFOs in New 
Mexico and does not include detailed infrastructure labels. 
METER-ML~\cite{zhu2022meter}, though broader in geographic scope 
(comprising four states), targets methane-related facilities and employs 
multi-resolution imagery~(NAIP and Sentinel-2).  Chugg~et~al.~\cite{CHUGG2021102463} target the growth of CAFO facilities
over time: they combine state-of-the-art building segmentation with 
change detection to flag unpermitted barn construction using time-series
satellite images by monitoring 1,513 known CAFO sites with Planet’s 3m 
imagery. A recent work \cite{saha2025machine} uses a random forest model on
publicly available socio-environmental
data—not satellite imagery—to map AFOs. Key predictors included canopy cover, land surface 
temperatures, and phosphorus levels using training data on~18 US states. 
In contrast, our CAFOSat dataset spans 20 U.S. states and provides detailed annotations of multiple CAFO types (swine, poultry, cattle, dairy, mixed) along with associated infrastructure, supporting fine-grained and interpretable modeling at national scale (see Table~\ref{tab:cafo-dataset-comparison}).

\textbf{Weakly Supervised Labeling for inaccurate geolocations.}
Across the above-mentioned datasets, automated patching may omit key visual
elements such as barns, manure ponds, and grazing areas, particularly when
facility coordinates are imprecise or resolution is limited. Such omissions
can pose challenges for downstream tasks requiring fine-grained 
infrastructure recognition. Existing CAFO detection works
acknowledge this problem. Handan-Nader~et~al.~\cite{handan2019deep} 
consider a similar approach as our work based on class-activation 
maps\cite{selvaraju2017grad}, but only to recenter the predicted infrastructure, followed by 
manual validation of a subset of predictions. 
Building on this direction, several studies propose refinements through CAM-based localization and progressive feature learning. 
Comprehensive reviews~\cite{li2022weakreview} outline the broader landscape of weak supervision, highlighting issues related to inexact, incomplete, and inaccurate labels. While these methods provide valuable foundations for learning under spatial uncertainty, these approaches typically assume that the weak labels are still reasonably well aligned. In contrast, our setting involves noisy point annotations that may not even intersect the object of interest, requiring a more robust strategy for localization.

\textbf{Prompt-Guided Synthetic Image Augmentation in Remote Sensing.}
Conventional data augmentation in remote sensing—such as geometric or spectral transformations~\cite{zhu2017deep}—does not capture the semantic complexity needed for fine-grained tasks like CAFO infrastructure analysis. To improve data diversity and address class imbalance, recent work has explored using generative models and vision-language prompts to synthetically expand underrepresented classes~\cite{chen2023diffusionaug, wang2023promptdiffusion}. Vision-language models (e.g., GroundingDINO~\cite{liu2023groundingdino}) enable open-vocabulary object detection via natural language, while diffusion-based inpainting~\cite{rombach2022high} supports realistic structure editing. Moreover, DiffusionSat~\cite{khanna2023diffusionsat} and GeoSynth~\cite{sastry2024geosynth} demonstrate that diffusion models can generate high-resolution satellite imagery conditioned on text, metadata, and spatial context, while more task-specific studies have explored prompt-driven augmentation for Earth observation datasets~\cite{de2024data} and synthetic generation for rare objects~\cite{nguyen2024generating}. These works motivate the use of generative models in remote sensing, but our goal differs: rather than generating fully synthetic scenes, we apply targeted, infrastructure-aware inpainting to create label-preserving CAFO variations that increase structural diversity while maintaining semantic validity.


%% file: sec/3_data.tex
\section{Data Processing Pipeline}
\label{sec:data_process}
\begin{table*}[t]
\footnotesize
\centering
\caption{CAFOSat dataset composition by category and split, with infrastructure-level 
annotation counts. The \texttt{Verified}-Set is verified benchmark spanning 
states unseen during training; the Held-Out Set is the in-distribution holdout 
evaluation set. Infrastructure counts reflect manually annotated positive samples 
in the full dataset. Horses, Sheep/Goats, and Other/Unknown categories are retained 
in the released dataset but excluded from benchmarking due to insufficient or 
ambiguous sample sizes. $^\dagger$Full dataset (39,212) + synthetically augmented 
samples (6,454) $\approx$ 45,000 total patches reported in the abstract.}
\label{tab:dataset_stats}
\resizebox{\textwidth}{!}{%
\begin{tabular}{lrrrr|rrrr}
\toprule
 & \multicolumn{4}{c|}{\textbf{Split Counts}} 
 & \multicolumn{4}{c}{\textbf{Infrastructure Annotations}} \\
\cmidrule(lr){2-5} \cmidrule(lr){6-9}
\textbf{Category} 
  & \textbf{Train} 
  & \textbf{\texttt{Verified}-Set} 
  & \textbf{Held-Out Set} 
  & \textbf{Total}$^\dagger$
  & \textbf{Barn} 
  & \textbf{Manure Pond} 
  & \textbf{Grazing Area} 
  & \textbf{Other Infra.} \\
\midrule
Negative      & 14,954 & 1,570 & 3,116 & 20,771 & ---   & ---   & ---   & ---   \\
Swine         &  6,330 &   972 & 1,320 & 11,220 & 1,150 &   548 &   400 &   889 \\
Poultry       &  1,546 & 1,239 &   326 &  3,576 & 1,033 &   298 &   528 &   458 \\
Dairy         &    484 &   405 &   100 &  1,643 &   359 &   182 &   135 &   250 \\
Beef          &  1,092 &   281 &   232 &  2,002 &   257 &   101 &    94 &   207 \\
\midrule
\textbf{Total} 
  & \textbf{24,406} 
  & \textbf{4,467} 
  & \textbf{5,094} 
  & \textbf{39,212} 
  & \textbf{2,799} 
  & \textbf{1,129} 
  & \textbf{1,157} 
  & \textbf{1,804} \\
\bottomrule
\end{tabular}}
\end{table*}
A detailed description of all data sources used to construct CAFOSat, 
including satellite imagery, CAFO location inventories, and land use 
masks, is provided in the supplementary material 
(Section~\ref{sec:supp_datasources}).

\noindent\textbf{CAFOSat Overview.}
The pipeline produces CAFOSat, a dataset of $\sim$45,000 NAIP image patches
(833$\times$833\,px, 0.6\,m resolution) spanning 20 U.S. states, plus 20,771
curated negatives. The full release covers six livestock categories (Swine,
Poultry, Dairy, Beef, Horses, Sheep/Goats); benchmarking experiments use the
four major types (Swine, Poultry, Dairy, Beef) alongside the Negative class,
as Horses and Sheep/Goats are excluded due to insufficient sample sizes
(24 and 21 samples respectively). Each sample includes: (i)~a
facility-level classification label (Swine, Poultry, Dairy, Beef, Horses,
Sheep/Goats, or Negative); (ii)~infrastructure-level annotations (barn count,
manure pond count, grazing area and other structure presence) for 4,513 manually
verified patches (Table~\ref{tab:dataset_stats}); (iii)~geospatial metadata
including original and refined coordinates, bounding boxes, and state;
(iv)~augmentation metadata for 6,454 synthetic patches; and (v)~six pre-defined
train/val/test splits for reproducible benchmarking. The dataset, data loaders,
and processing scripts are publicly
available. Full details of all provided
fields and split configurations are described in the supplement
(Section~\ref{sup:data_process}).

An overview of the data processing pipeline is provided in
Figure~\ref{fig:data_process}. We follow five steps to generate CAFOSat:

\noindent\textbf{Standardized Training Data Creation.} To construct a
high-quality training dataset, we start by selecting all CAFOs from the
CAFOMaps~\cite{cafomaps} dataset, which has a diverse set of livestock operations
(e.g., poultry, beef, dairy, and swine) to ensure sufficient sample diversity.
We follow standard remote sensing practices by first geo-locating all 4513
verified CAFO coordinates and mapping them to the corresponding NAIP satellite
imagery. For each location, we extract a fixed-size $833\times833$ pixel patch
centered on the CAFO. The patch dimensions are carefully selected based on domain
knowledge (i.e., CAFO infrastructure can span up to 1,000--1,500 meters in
length), ensuring that the full extent of facilities such as barns, manure
lagoons, and feedlots is captured. All generated patches are then manually
verified and annotated using binary labels (CAFO or non-CAFO).

\noindent\textbf{AI Annotator.} To construct our initial \textit{Standard
Training Set}, we begin with labeled CAFO samples from
RegLab~\cite{handan2019deep}, which includes a small number of manually verified
poultry and swine facilities. Using this data, we train a ResNet-based binary
classification model, which we then employ as an AI annotator to identify likely
CAFO patches from our collected CAFOMaps~\cite{cafomaps} dataset --- selected for
its broader diversity of CAFO types. To reduce manual effort and improve
efficiency, we use the AI annotator to pre-filter candidate patches, focusing
only on those where the model predicts CAFO presence with high confidence
(softmax probability $\geq 0.8$). This strategy significantly narrows the
verification workload while maintaining label quality. The model is applied
specifically to poultry and swine patches, and all 3,286 positively predicted
samples (1,615 Swine + 1,671 Poultry) are manually verified to ensure high label
quality. This verification is essential, as the original location data is noisy,
and our goal is to build a small, strongly annotated training set. To further
increase class diversity, we manually verify all dairy and beef patches, obtaining
an additional 383 dairy and 209 beef CAFO samples. These verified samples,
combined with carefully curated negative examples, comprise our \textbf{\textit{Standard
Training Set}}, which is used throughout the rest of our experiments. Note that
while the Iowa data is used for candidate selection and verification, it is
excluded from training due to its differing image resolution.

\noindent\textbf{Refining Weakly Annotated Data.} Given that CAFO location
annotations are often weak, we propose a refinement algorithm that transforms
these coarse coordinates into spatially accurate point annotations using
model-guided attention~(GradCAM) and geometric clustering. Our process is shown
in Algorithm~\ref{alg:refine}. Starting from a weak coordinate annotation
$\ccenter$, we extract five patches of shape $833\times833$ pixels: one centered
at~$\ccenter$ and four diagonally offset by~416 pixels to ensure local spatial
coverage. Each patch is evaluated using the AI annotator created in the previous
step. Patches predicted as CAFOs are then processed with GradCAM to generate
class-specific attention heatmaps. These heatmaps are thresholded at level~$\tau$
to isolate high-confidence activation regions, which are converted into polygonal
shapes via connected-component analysis. GradCAM heatmaps are computed using the
\textit{torchcam} library~\cite{fernandez2020torchcam}. To localize the dominant CAFO
structure, we apply single-linkage clustering to the extracted polygons using a
minimum inter-polygon distance threshold~$\dmerge$, which groups spatially
contiguous regions that likely correspond to a single CAFO. The final refined
coordinate is computed as the centroid of the merged polygon from the largest
cluster~(based on polygon counts). Patches associated with smaller clusters are
discarded, as they likely correspond to spurious activations or unrelated
structures rather than the primary CAFO facility.

\begin{algorithm}[t]
\footnotesize
\caption{CAFO Coordinate Refinement via Classification and GradCAM Clustering}
\label{alg:refine}
\KwIn{Image $I$; center coordinate $c_{\text{center}} \in \mathbb{R}^2$;
classifier $\mathcal{A}$; GradCAM threshold~$\tau$; clustering distance $\dmerge$}
\KwOut{Refined coordinate $c^* \in \mathbb{R}^2$}
\vspace{1mm}
$r \gets 416$ \;
$c_0 \gets c_{\text{center}}$ \;
$c_1 \gets c_0 + (r, r)$, $c_2 \gets c_0 + (r, -r)$, $c_3 \gets c_0 + (-r, r)$,
$c_4 \gets c_0 + (-r, -r)$ \;
Extract patches $P_i$ of size $833 \times 833$ pixels centered at $c_i$ for
$i = 0,\ldots,4$ \;
$\mathcal{P}_{\text{CAFO}} \gets \{P_i \mid \mathcal{A}(P_i) = 1\}$ \;
$\mathcal{G} \gets \emptyset$ \;
\ForEach{$P \in \mathcal{P}_{\text{CAFO}}$}{
    $H \gets \text{GradCAM}(P)$ \;
    $M \gets \{(x, y) \mid H(x, y) \geq \tau\}$ \;
    Extract connected components $\{R_k\}$ from $M$ \;
    Convert each $R_k$ to minimum bounding polygon $G_k$ and update
    $\mathcal{G} \gets \mathcal{G} \cup \{G_k\}$ \;
}
Perform single-linkage clustering on $\mathcal{G}$:
\(
\text{dist}(G_i, G_j) = \min_{x \in G_i,\; y \in G_j} \|x - y\|_2
\)\;
Let $\mathcal{G}^* \subseteq \mathcal{G}$ be the largest resulting cluster \;
$G_{\text{merged}} \gets \bigcup \mathcal{G}^*$ \;
$c^* \gets \text{Centroid}(G_{\text{merged}})$ \;
\Return{$c^*$} \;
\end{algorithm}

\noindent\textbf{Manual Verification \& Annotation.} To assess annotation
quality, we manually verified a total of 4,513 samples. The \texttt{Verified}
set was constructed using a stratified sampling strategy: we included all 2,344
samples labeled as negative by the AI annotator, and additionally sampled 2,169
instances uniformly at random from the remaining pool. For each CAFO patch, we
provide infrastructure-level annotations: we manually count the number of visible
barns and manure ponds, and record the presence or absence of grazing areas and
other non-barn structures.

\noindent\textbf{Negative Sample Generation.} We employ a stratified negative
sampling strategy guided by land use data to generate non-CAFO training samples.
Specifically, we use 30-meter resolution land cover maps to identify land types
where CAFOs are plausibly located. Across all studied states, we select eight
relevant land use classes: pasture, grassland, developed, semi-developed,
cropland, barren land, shrubland, and open space/water. From each land use
category, we randomly sample 3,000 candidate coordinates, resulting in a diverse
pool of negative samples stratified by land cover type. To ensure that sampled
points do not overlap with nearby CAFO infrastructures, we apply a 1.5-kilometer
exclusion radius around all known CAFO locations (any candidate point within this
buffer was discarded). The remaining coordinates are used to extract fixed-size
image patches~($833\times833$ pixels) centered on each location, ensuring all
patches fall within valid image bounds. This process yields a total of 20,771
high-confidence negative samples for robust classifier training.

\noindent\textbf{Prompt-Guided Data Augmentation.} To improve model
generalization in CAFO infrastructure analysis, we develop a prompt-guided
augmentation pipeline that removes visually identifiable structures, such as
barns, manure ponds, and other supporting facilities, and replaces them with
semantically plausible non-infrastructure content. This results in
label-preserving image variants that diversify structural configurations while
retaining the associated CAFO types (e.g., swine, dairy, poultry) as defined in
the metadata. In total, 6,454 unique patches were augmented; individual patches
may contain multiple infrastructure types, yielding 3,921 barn, 1,344 manure
pond, and 2,089 other infrastructure removal operations across the augmented set
(see Figure~\ref{fig:augment_examples} in Supplement).

First, we perform \textbf{infrastructure detection via vision-language prompts},
where the objective is to localize and identify physical structures within
satellite imagery. Using GroundingDINO~\cite{liu2023groundingdino}, a
vision-language object detector, we detect infrastructure based on natural
language prompts that encode high-level visual priors observed in overhead
imagery. We carefully designed 25 prompts that capture structural characteristics
of CAFOs across livestock types; details are in the supplement. The second part
is \textbf{inpainting for structure removal}, where we generate binary masks from
subsets of predicted bounding boxes~(up to five per image) and use these to guide
the removal of detected infrastructure. The RGB image and corresponding mask are
passed to a Stable Diffusion Inpainting model~\cite{rombach2022ldm},
conditioned on non-infrastructure prompts (e.g., ``grassland'', ``a small water
pool'', ``cluster of trees''). The model synthesizes contextually consistent
content to fill the masked regions. To preserve the validity of the CAFO type
label, we remove infrastructure only when multiple instances are present within a
patch --- for example, if ten barns are detected, a subset (e.g., two or three)
are removed to ensure the structural identity of the facility remains intact. This
process generates the \texttt{Augmented Set}. Detailed metadata is provided for
each sample to ensure full traceability and label integrity across all
synthetically generated patches.

Our prompt-guided augmentation strategy builds on a growing body of work
applying generative models to expand remote sensing training data. Conventional
augmentation approaches such as geometric and spectral
transformations~\cite{zhu2017deep} do not capture the semantic variability needed
for fine-grained infrastructure recognition. Recent work has explored
diffusion-based inpainting and vision-language guidance to synthesize
semantically consistent remote sensing
imagery~\cite{chen2023diffusionaug, wang2023promptdiffusion}, including
satellite-image-specific generative models such as DiffusionSat~\cite{khanna2023diffusionsat}
and GeoSynth~\cite{sastry2024geosynth}. Our approach extends this direction to
agricultural infrastructure, using GroundingDINO for open-vocabulary
localization and Stable Diffusion for structure-aware inpainting, while
preserving CAFO type labels through a multi-instance removal constraint.

%% file: sec/4_experiments.tex
\section{Analysis and Benchmarking}
\label{sec:benchmarking}
\begin{table*}[t]
\footnotesize
\caption{Performance comparison of different models on two evaluation sets.
The \texttt{Verified}-Set (4,513 samples) spans out-of-distribution states unseen during
training; the Held-Out Set (5,103 samples) is an in-distribution random
holdout. F1 and mAP are macro-averaged. Per-class F1 scores are reported
for the four major CAFO types.}
\label{tab:valset_comparison}
\centering
\resizebox{\textwidth}{!}{%
\begin{tabular}{lrrr|rrrr|rrr|rrrr}
\toprule
& \multicolumn{7}{c|}{\textbf{\texttt{Verified}-Set}}
& \multicolumn{7}{c}{\textbf{Held-Out Set}} \\
\cmidrule(lr){2-8} \cmidrule(lr){9-15}
\textbf{Model}
  & \textbf{Acc.} & \textbf{F1} & \textbf{mAP}
  & \textbf{Swine} & \textbf{Poultry} & \textbf{Dairy} & \textbf{Beef}
  & \textbf{Acc.} & \textbf{F1} & \textbf{mAP}
  & \textbf{Swine} & \textbf{Poultry} & \textbf{Dairy} & \textbf{Beef} \\
\midrule
ResNet18
  & 0.596 & 0.383 & 0.401 & 0.579 & 0.838 & 0.405 & 0.574
  & 0.939 & 0.481 & 0.511 & 0.930 & 0.851 & 0.391 & 0.695 \\
ResNet50
  & 0.601 & 0.399 & \textbf{0.444} & 0.578 & 0.819 & 0.591 & 0.608
  & 0.935 & 0.482 & 0.513 & 0.923 & 0.830 & 0.505 & 0.613 \\
ViT-B/16
  & 0.596 & 0.377 & 0.374 & 0.581 & 0.826 & 0.545 & 0.437
  & 0.934 & 0.492 & 0.528 & 0.919 & 0.859 & 0.569 & 0.609 \\
Swin-B
  & 0.630 & \textbf{0.421} & 0.443 & \textbf{0.600} & 0.853
  & \textbf{0.655} & \textbf{0.655}
  & \textbf{0.952} & 0.527 & 0.559 & \textbf{0.946} & 0.873
  & \textbf{0.670} & \textbf{0.736} \\
ConvNeXt
  & 0.626 & 0.400 & 0.412 & 0.592 & 0.852 & 0.626 & 0.633
  & 0.945 & \textbf{0.544} & \textbf{0.593} & 0.938 & 0.861 & 0.622 & 0.691 \\
EfficientNet
  & \textbf{0.636} & 0.405 & 0.431 & 0.599 & \textbf{0.867} & 0.634 & 0.640
  & 0.949 & 0.517 & 0.547 & \textbf{0.946} & \textbf{0.890} & 0.581
  & 0.729 \\
CLIP
  & 0.492 & 0.292 & 0.298 & 0.503 & 0.675 & 0.441 & 0.308
  & 0.878 & 0.415 & 0.420 & 0.855 & 0.706 & 0.352 & 0.445 \\
RemoteCLIP
  & 0.560 & 0.317 & 0.342 & 0.544 & 0.767 & 0.461 & 0.327
  & 0.912 & 0.453 & 0.486 & 0.898 & 0.764 & 0.488 & 0.497 \\
DINOv2
  & 0.602 & 0.364 & 0.371 & 0.569 & 0.835 & 0.524 & 0.530
  & 0.921 & 0.469 & 0.483 & 0.910 & 0.811 & 0.468 & 0.587 \\
\bottomrule
\end{tabular}}
\end{table*}
We evaluate CAFOSat along four dimensions: ($i$) the effect of annotation
refinement on classification performance, ($ii$) cross-dataset transfer between
CAFOSat and prior CAFO datasets, ($iii$) multiclass benchmark performance on
two standardized CAFOSat evaluation splits, and ($iv$) the impact of synthetic
augmentation on model robustness.

\noindent\textbf{Evaluation Sets.} We define two complementary evaluation sets.
($i$)~\textbf{\texttt{Verified}-Set:} 4,513 manually verified samples, including unseen states during training (AR, CA, FL, GA), providing a controlled benchmark for geographic generalization.
($ii$)~\textbf{Held-Out Set:} 5,103 samples held out from in-distribution
states, serving as a standard random holdout evaluation. Model details and
metrics are provided in the supplement.

\noindent\textbf{Impact of Annotation Quality on Performance.}
We study whether improving coordinate alignment (via manual refinement)
increases CAFO classification performance. We train binary classifiers on
three sources---RegLab~\cite{handan2019deep}, METER-ML~\cite{zhu2022meter},
and our \textit{Standard Training Set} (Sec.~\ref{sec:data_process})---and
evaluate on CAFOSat CAFO-only patches extracted using either the original
(survey-reported) coordinates or the refined coordinates. We restrict this
evaluation to CAFO patches because refinement is applied to known CAFO
locations; a well-centered patch should contain visible CAFO infrastructure
and thus be classified as CAFO. Table~\ref{tab:cafo-refined-performance}
compares performance on weak vs.\ refined patch extractions. Across models,
refined patches consistently yield higher accuracy, indicating that better
spatial alignment increases the visibility of discriminative CAFO structures
and improves recognition.

We further assess cross-dataset generalization to position CAFOSat as both
a training resource and a benchmark. Table~\ref{tab:cafo_comparison_T1}
reports models trained on METER-ML and RegLab and evaluated on the full
CAFOSat dataset (CAFO and non-CAFO). METER-ML trained models transfer reasonably well,
whereas RegLab trained models transfer is weaker, highlighting domain gaps (i.e., state-wise variation) and the need for
representative supervision. Conversely, Table~\ref{tab:cafo_comparison_T2}
shows that models trained on CAFOSat generalize well to the \texttt{Verified}-Set,
METER-ML, and RegLab; transformer-based models (e.g., ViT-B/16) achieve
strong accuracy and F1 on the \texttt{Verified}-Set. Overall, these results demonstrate
that CAFOSat provides high-quality supervision, benefits from refined
annotations, and supports robust cross-dataset evaluation.

\begin{table*}[h]
\footnotesize
\caption{Impact of annotation quality on CAFO detection performance.
Binary classifiers are trained on three source datasets (RegLab, METER-ML,
and our Standard Training Set) and evaluated on CAFOSat CAFO-only patches
extracted using either weakly annotated (Weak.) survey coordinates or
spatially refined (Ref.) coordinates. Higher scores indicate better
recognition of visible CAFO infrastructure.}
\label{tab:cafo-refined-performance}
\centering
\begin{tabular}{lcccccccc}
\toprule
\textbf{Train Dataset} & \multicolumn{2}{c}{\textbf{ResNet18}}
                 & \multicolumn{2}{c}{\textbf{ResNet50}}
                 & \multicolumn{2}{c}{\textbf{ViT-B/16}}
                 & \multicolumn{2}{c}{\textbf{Swin-B}} \\
\cmidrule(lr){2-3} \cmidrule(lr){4-5} \cmidrule(lr){6-7} \cmidrule(lr){8-9}
 & Weak. & Ref. & Weak. & Ref. & Weak. & Ref. & Weak. & Ref. \\
\midrule
RegLab~\cite{handan2019deep}
  & 0.357 & 0.375 & 0.070 & 0.071 & 0.552 & 0.601 & 0.521 & 0.547 \\
METER-ML~\cite{zhu2022meter}
  & 0.595 & 0.659 & 0.441 & 0.507 & 0.718 & 0.766 & 0.638 & 0.660 \\
\textit{Standard Training Set}
  & \textbf{0.767} & \textbf{0.883} & \textbf{0.870} & \textbf{0.931}
  & \textbf{0.791} & \textbf{0.894} & \textbf{0.891} & \textbf{0.979} \\
\bottomrule
\end{tabular}
\end{table*}

\begin{table}[t]
\footnotesize
\caption{Evaluating external dataset generalization on the CAFOSat benchmark.
Models trained on METER-ML or RegLab are evaluated on the full CAFOSat dataset
(binary CAFO vs.\ non-CAFO).}
\label{tab:cafo_comparison_T1}
\centering
\begin{tabular}{p{2.2cm}p{1.4cm}p{1cm}p{1cm}p{1cm}}
\toprule
\textbf{Train Dataset} & \textbf{Model} & \textbf{Acc.} & \textbf{F1}
  & \textbf{CAFO Acc.} \\
\midrule
\multirow{2}{*}{METER-ML~\cite{zhu2022meter}}
  & Swin-B   & 0.823 & 0.809 & 0.659 \\
  & ViT-B/16 & \textbf{0.869} & \textbf{0.862} & \textbf{0.766} \\
\midrule
\multirow{2}{*}{RegLab~\cite{handan2019deep}}
  & Swin-B   & 0.739 & 0.686 & 0.394 \\
  & ViT-B/16 & \textbf{0.786} & \textbf{0.768} & \textbf{0.601} \\
\bottomrule
\end{tabular}
\end{table}

\begin{table}[h]
\footnotesize
\caption{Benchmarking CAFOSat-trained models on the \texttt{Verified}-Set and external
datasets for binary CAFO vs Non-CAFO classification. R denotes macro recall; CAFO Acc.\ is accuracy on positive
CAFO samples only.}
\label{tab:cafo_comparison_T2}
\begin{tabular}{p{1.8cm}p{1.2cm}p{.6cm}p{.6cm}p{.6cm}p{.8cm}}
\toprule
\textbf{Test Dataset} & \textbf{Model} & \textbf{Acc.} & \textbf{F1}
  & \textbf{R} & \textbf{CAFO Acc.} \\
\midrule
\multirow{4}{*}{\texttt{\texttt{Verified}-Set}}
  & ResNet18 & 0.714 & 0.631 & 0.627 & \textbf{0.911} \\
  & ResNet50 & 0.701 & 0.609 & 0.610 & \textbf{0.911} \\
  & ViT-B/16      & \textbf{0.928} & \textbf{0.925} & \textbf{0.921} & 0.878 \\
  & Swin-B   & 0.702 & 0.614 & 0.614 & 0.902 \\
\midrule
\multirow{4}{*}{METER-ML~\cite{zhu2022meter}}
  & ResNet18 & 0.831 & 0.817 & 0.808 & 0.658 \\
  & ResNet50 & \textbf{0.849} & \textbf{0.836} & \textbf{0.825} & \textbf{0.676} \\
  & ViT-B/16      & 0.761 & 0.719 & 0.718 & 0.441 \\
  & Swin-B   & 0.810 & 0.788 & 0.779 & 0.580 \\
\midrule
\multirow{4}{*}{RegLab~\cite{handan2019deep}}
  & ResNet18 & 0.944 & 0.822 & 0.777 & 0.567 \\
  & ResNet50 & 0.946 & 0.813 & 0.750 & 0.504 \\
  & ViT-B/16      & 0.914 & 0.619 & 0.582 & 0.165 \\
  & \textbf{Swin-B} & \textbf{0.966} & \textbf{0.903} & \textbf{0.880}
    & \textbf{0.772} \\
\bottomrule
\end{tabular}
\end{table}


\begin{table}[t]
\footnotesize
\caption{Performance across two training setups evaluated on the
\textbf{Held-Out Set}. \textit{Annotated}: manually verified samples
only; \textit{Augmented}: synthetic inpainted samples only.}
\label{tab:valset_swin_conv_comparison}
\centering
\begin{tabular}{p{1.4cm}p{1.2cm}p{1.7cm}p{1.7cm}}
\toprule
\textbf{Model} & \textbf{Type} & \textbf{F1} & \textbf{mAP} \\
\midrule
\multirow{2}{*}{\textit{Swin-B}}
  & Annotated   & 0.770 & 0.495 \\
  & Augmented & 0.764 (0.6\%$\downarrow$) & 0.473 (2.2\% $\downarrow$) \\
\midrule
\multirow{2}{*}{\textit{ConvNeXt}}
  & Annotated   & 0.835 & 0.500 \\
  & Augmented & 0.799 (3.6\% $\downarrow$) & 0.480 (2.0\% $\downarrow$) \\
\bottomrule
\end{tabular}
\end{table}

\noindent\textbf{Benchmarking.} To benchmark the effectiveness of deep learning
models for multiclass CAFO classification, we evaluated a suite of nine
diverse architectures: ResNet18/50~\cite{he2016deep},
EfficientNet~\cite{tan2019efficientnet}, ConvNeXt~\cite{liu2022convnet},
ViT-B/16~\cite{dosovitskiy2020vit}, Swin-B~\cite{liu2021swin},
DINOv2~\cite{oquab2023dinov2}, CLIP~\cite{radford2021clip}, and
RemoteCLIP~\cite{mallya2022remoteclip}. These models span a range of design
paradigms---from classical CNNs to self-attention-based transformers and
multimodal pretrained systems---providing a comprehensive view of current
capabilities. All models were trained on our curated CAFO dataset and
assessed across two standardized evaluation sets: (i)~\textit{\texttt{Verified}-Set} and
(ii)~\textit{Held-Out Set}. The results in Table~\ref{tab:valset_comparison}
show that transformer-based models (e.g., Swin-B and EfficientNet) generally
outperform CNN-based baselines such as ResNet, with notably stronger per-class
F1 and mean Average Precision (mAP). Across both sets, Poultry is the
easiest class to recognize while Dairy is the hardest, due to visual
similarity with Beef CAFO structures. More results are
reported in Table~\ref{tab:cafo_classwise_f1} and Figure~\ref{fig:conf}
in the supplementary material.

\noindent\textbf{Effectiveness of Synthetic Augmentation.}
To quantify how much synthetic CAFO imagery can substitute for, and complement, real annotations during training, we compare models trained on real-only (samples from \texttt{Verified}-set) and synthetic-only (samples from \texttt{Augmented} dataset). Note that each
training set included 20\% negative samples from CAFOSat for class balance. We select best performing vision transformer model (Swin-B) and convolutional model (ConvNeXt)---top-performing models from prior
benchmarking. As shown in Table~\ref{tab:valset_swin_conv_comparison}, we notice that model trained on augmented dataset produce similar performance (with a drop of only 0.6\% F1 and 2.2\% mAP) as model trained on verified annotated dataset. It is worth noting that performance gap for transformer based model are lower than CNN based model. These results underscore the complementary value of augmentation, particularly for transformer-based
models in data-scarce or imbalanced contexts.

%% file: sec/5_Discussion.tex
\section{Conclusion}
CAFOSat addresses a key obstacle in scalable CAFO mapping: public inventories exist at scale, but their geolocations are often noisy, and this misalignment substantially degrades learning from overhead imagery. Our results show that refining weak point labels is a first-order driver of performance, often rivaling the effect of changing model families, by re-centering patches on the visually salient infrastructure. Across architectures, transformer-based and pretrained models generally achieve the best overall accuracy-especially in CAFO recall and mAP-suggesting that global context and layout cues are critical for distinguishing CAFOs from visually similar agricultural facilities. We also find that synthetic augmentation can improve robustness by diversifying the training distribution, though benefits are not uniform, motivating adaptive filtering/weighting of synthetic samples. Limitations include incomplete geographic coverage and potential reporting biases in source inventories, residual confusion with hard negatives, and the simplifications inherent to patch-based classification; future work should extend CAFOSat toward detection/segmentation, temporal modeling, and uncertainty-aware screening for large-area monitoring. CAFOSat inherits biases from public inventories, imagery, and annotation procedures, leading to potential regional coverage gaps, outdated records, and uneven label quality. Additionally, its U.S.-centric scope and underrepresentation of rare facility configurations may limit generalization to other geographies and atypical infrastructure layouts.  Moreover, publicly available ML-ready geospatial datasets and regulatory definitions are often less standardized internationally, limiting direct comparison across countries. In future, we will work on foundation for extending infrastructure-aware livestock mapping to broader international settings.

\section*{Acknowledgment}
{This material is based upon work supported by the AI Research Institutes program supported by NSF and USDA-NIFA under the AI Institute: Agricultural AI for Transforming Workforce and Decision Support (AgAID) award No.~2021-67021-35344. This work was
partially supported by University of Virginia Strategic Investment Fund award number SIF160.}


%% file: sec/X_suppl.tex
\clearpage
\appendix
\setcounter{page}{1}
\maketitlesupplementary


\setcounter{figure}{0}
\setcounter{table}{0}
\renewcommand{\thefigure}{S\arabic{figure}}
\renewcommand{\thetable}{S\arabic{table}}

\bigskip
\baselineskip = 1.10\normalbaselineskip

\begin{figure*}
    \centering
    \includegraphics[width=.8\linewidth]{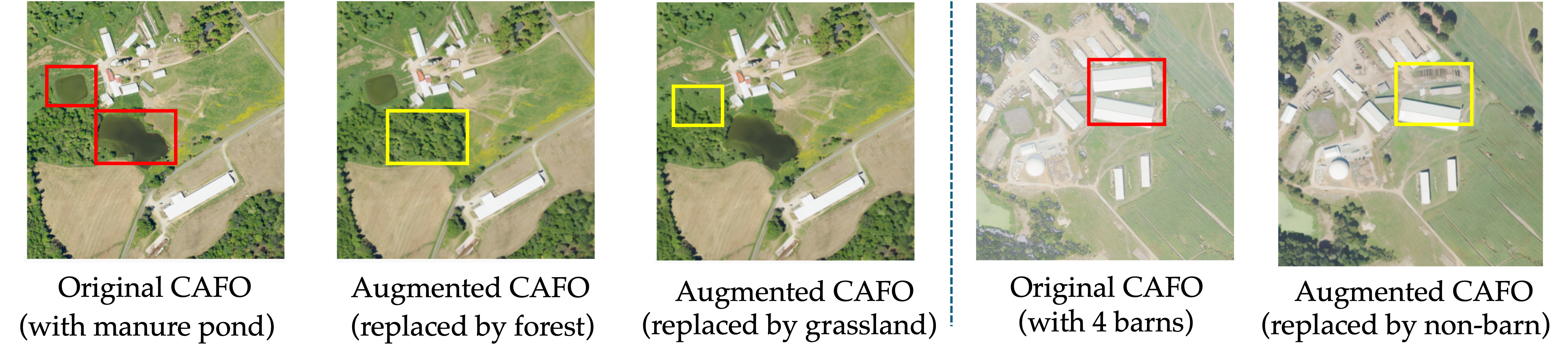}
    \caption{Examples from our augmented dataset. In the first example (left),
    the augmented images replace the manure pond with forest and grassland; in
    the second example (right), one barn is replaced with non-barn infrastructure.
    In both cases, the structural definition of the CAFO remains intact.}
    \label{fig:augment_examples}
\end{figure*}

\section{CAFOSat Overview.}
\label{sup:data_process}
An overview of the data processing pipeline is provided in 
Figure~\ref{fig:data_process}.
The pipeline described in this section produces CAFOSat, a comprehensive, 
ML-ready dataset for CAFO detection and infrastructure mapping across the 
contiguous United States. The full dataset comprises \textbf{39,257 base image 
patches} (833$\times$833 pixels, 0.6\,m resolution) from 2023 NAIP imagery, 
spanning 20 U.S. states and six livestock categories (Swine, Poultry, Dairy, 
Beef, Horses, Sheep/Goats), plus 20,771 curated negative samples. An additional 
6,454 synthetically augmented patches are included, bringing the total to 
approximately 45,000 patches. For each sample, CAFOSat provides:
\begin{itemize}[leftmargin=*, noitemsep]
    \item \textbf{Image patches:} High-resolution NAIP aerial imagery 
    (833$\times$833 px, 0.6\,m/px) centered on CAFO or non-CAFO locations, 
    organized by U.S. state.

    \item \textbf{Classification labels:} Facility-level CAFO type labels 
    (Negative, Swine, Poultry, Dairy, Beef, Horses, Sheep/Goats) with 
    human-verified and AI-annotated variants indicated via a 
    \texttt{verified\_label} flag.

    \item \textbf{Infrastructure-level annotations:} Binary flags for barn 
    presence, manure pond presence, grazing area presence, and other 
    infrastructure for 4,513 manually verified patches 
    (see Table~\ref{tab:dataset_stats} for per-class counts).

    \item \textbf{Bounding box annotations:} Geospatial bounding boxes 
    (\texttt{geom\_bbox}) and polygon geometries (\texttt{geometry}) with 
    associated coordinate reference systems for each patch.

    \item \textbf{Geospatial metadata:} Original weak coordinates 
    (\texttt{weak\_x}, \texttt{weak\_y}) alongside refined coordinates 
    (\texttt{refined\_x}, \texttt{refined\_y}), U.S. state, spatial 
    resolution, and unique facility identifiers (\texttt{CAFO\_UNIQUE\_ID}).

    \item \textbf{Augmentation metadata:} For synthetic patches, the 
    original source patch path, the inpainting prompt used, and the 
    \texttt{image\_type} flag distinguishing real from augmented samples.

    \item \textbf{Standardized splits:} Six pre-defined train/val/test 
    split configurations (Verified, Augmented, Merged, Set~1, Set~2, 
    All-Training) provided as boolean flags in the master metadata file 
    (\texttt{CAFOSat.csv}), enabling reproducible benchmarking across 
    different experimental setups (see Table~\ref{tab:dataset_stats}).
\end{itemize}
The dataset is publicly available on HuggingFace\footnote{\url{https://huggingface.co/datasets/oishee3003/CAFOSat}} 
under a CC BY 4.0 license, with data loaders and processing scripts 
on GitHub\footnote{\url{https://github.com/oishee-hoque/CAFOSat}}. 
The remainder of this section details how each component is constructed.

\section{Data Sources}
\label{sec:supp_datasources}
Here, we provide a summary of data sources and acquisition. Details are in the supplement. 
Table~\ref{tab:cafo-sources} lists all data sources.

\subsection{Satelite Imagery}
We leverage aerial imagery from the National Agriculture Imagery Program (NAIP), accessed via the Microsoft Planetary Computer. For each geolocated CAFO point, the nearest cloud-free NAIP image is queried and downloaded using the STAC API. These images typically offer spatial resolutions between 60 cm and 1 m, depending on acquisition year and state. For this experimentation, we collected the most recent (2023) data for the studied states.

\begin{table*}[h]
\centering
\footnotesize
\caption{Summary of Core CAFO Data Sources Used in Livestock Detection.}
\label{tab:cafo-sources}
\begin{tabular}{p{3cm}p{8cm}p{4.5cm}}
\toprule
\textbf{Source} & \textbf{Description} & \textbf{Use} \\
\midrule
NAIP Imagery (2023) ~\cite{usda_naip} & High-resolution aerial imagery from the USDA National Agriculture Imagery Program & Visual input for CAFO patch extraction and model training \\ \hline
CAFOMaps \cite{cafomaps} & Multi-state (i.e., Alabama, Arkansas, Florida, Georgia, Louisiana, Mississippi, North Carolina, South Carolina, and Texas) labeled dataset containing 6,604 CAFOs with animal type annotations (e.g., poultry, swine, beef, dairy), curated by IOWA researchers & Ground-truth labels for training and validation of CAFO classification models \\\hline
State-CAFO Inventory \cite{INcafo,IAcafo,MDcafo,MIcafo,MNcafo,NYcafo,DEcafo} & Six independent data sources from official state-level CAFO registries 
corresponding to Indiana, Iowa, Maryland, Michigan, Minnesota, New York, and Delaware, curated from permit records, inspections, and nutrient management plans & Ground truth labels for ML-ready dataset \\\hline
Land Use Masks (NLCD) \cite{nlcd}& National Land Cover Database (NLCD) used for masking agricultural zones & Used to create negative samples \\
\bottomrule
\end{tabular}
\end{table*}


\subsection{CAFO-Dataset}
We collected CAFO location and corresponding cafo type primarily from two sources: ($i$) Department of Geographical and Sustainability Sciences of IOWA\footnote{\url{https://cafomaps.org/}} (denoted as IOWA-CAFO Inventory) and ($ii$) animal feeding operation report from states (denoted as State-CAFO Inventory). 

\textbf{IOWA-CAFO Inventory:} This data inventory aggregates CAFO facility data from state environmental agencies across nine southeastern U.S. states: Alabama, Arkansas, Florida, Georgia, Louisiana, Mississippi, North Carolina, South Carolina, and Texas. Data is collected from permit databases, nutrient management plans, and agency inspections. Each record includes geolocation, animal type (poultry, swine, beef, dairy), and manure management details. While some states (e.g., Georgia, North Carolina) provide structured digital datasets, others require manual extraction from reports. This dataset standardizes these sources to enable regional CAFO distribution and impact analysis.



\textbf{State-CAFO Inventory} Several U.S. states publish CAFO reports curated by official state agencies using permit records, inspection data, and self-reported nutrient management plans. Delaware’s CAFO report, maintained by the Socially Responsible Agricultural Project (SRAP), provides spatial boundaries and regulatory attributes for permitted operations. Indiana’s CAFO report, compiled by the Indiana Department of Environmental Management (IDEM), is publicly available via an ArcGIS map and contains metadata on facility type and permit status. Iowa’s inventory is provided by the Iowa Department of Natural Resources (Iowa DNR), offering geolocated animal feeding operations with supporting regulatory information. Maryland’s dataset is curated by the Maryland Department of the Environment (MDE), detailing registered CAFOs with geographic and operational metadata. Michigan’s CAFO data is managed by the Michigan Department of Environment, Great Lakes, and Energy (EGLE), covering operations across different time periods with detailed facility-level information. Additional inventories are available from the FracTracker Alliance and the New York Department of State (NYDOS), including permit identifiers and operator details across various temporal spans.

\subsection{Land Use Masks (NLCD)}
We utilize national-scale raster products to identify and contextualize  agricultural areas. The MRLC National Land Cover 
Database~(NLCD) offers 30m-resolution land cover 
classifications across 16 categories, including cultivated cropland, grassland, barren land, and pasture~(See Table~\ref{tab:cafo-sources}). This dataset is further used to generate stratified negative samples based on land cover types and spatial extent.

\section{Additional Details for Section~\ref{sec:data_process}}
\begin{figure}[h]
    \centering
    \includegraphics[width=1\linewidth]{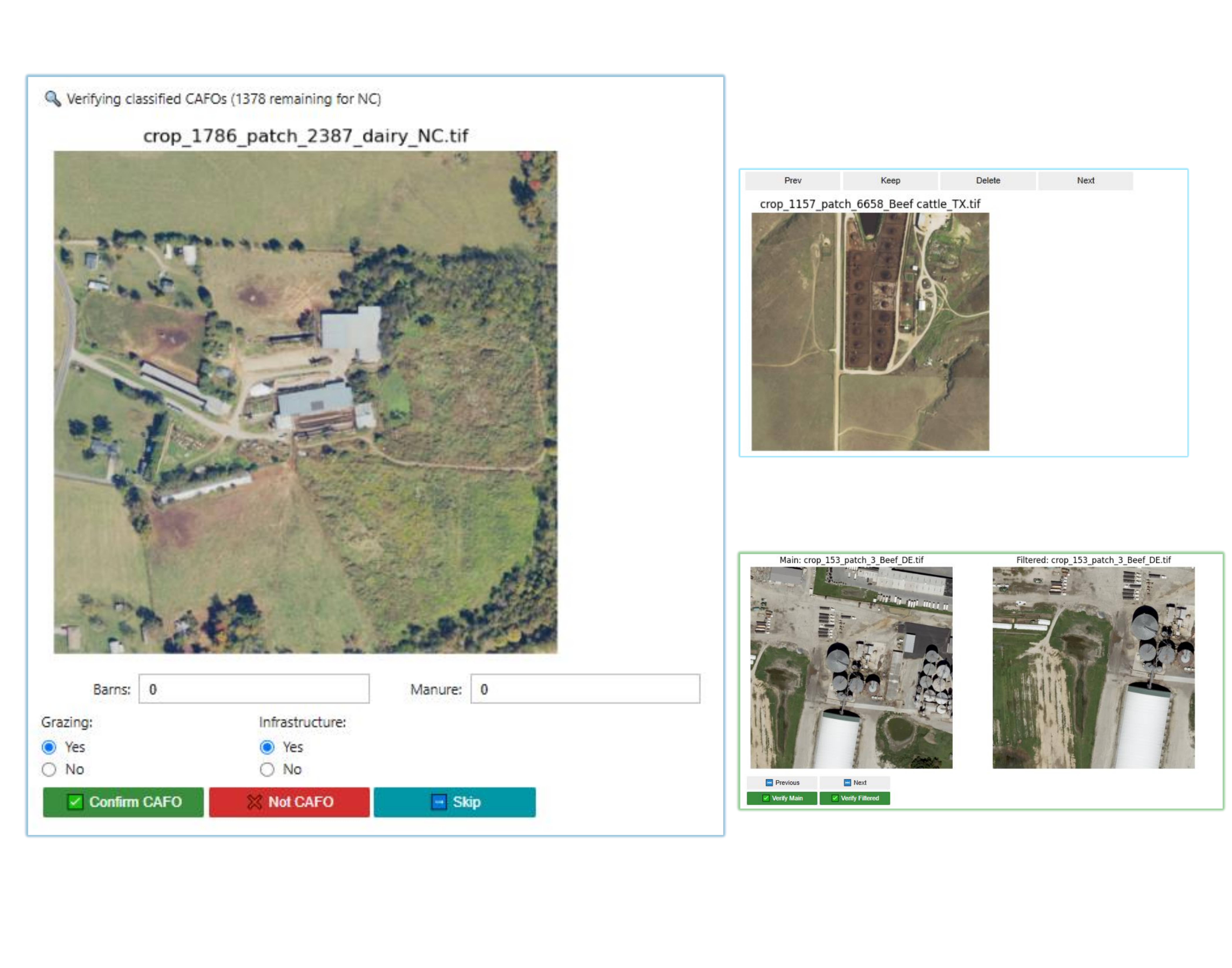}
    \caption{Several Manual Verification and Labeling Setup}
    \label{manual}
\end{figure}

\paragraph{Manual Verification Setup.}
Figure \ref{manual} visualizes the verification setup we used to manually verify and annotate the data.
\paragraph{Prompt Guided Data Augmentation} To improve model generalization in CAFO infrastructure analysis, we develop a prompt-guided augmentation pipeline that removes visually identifiable structures, such as barns, manure ponds, and other supporting facilities. We then replace them with semantically plausible non-infrastructure content. This results in label-preserving image variants that diversify structural configurations while retaining the associated CAFO types (e.g., swine, dairy, poultry) as defined in the metadata. In total, 6454 samples were augmented using the following prompt-guided procedure (3921 involving barns, 1344 manure ponds, and 2089 other infrastructure elements):

\begin{itemize}
    \item \textbf{Infrastructure Detection via Vision-Language Prompts}: In this step our objective is to localize and identify physical structures within satellite imagery. Using GroundingDINO, a vision-language object detector, we detect infrastructure based on natural language prompts that encode high-level visual priors observed in overhead imagery. We carefully designed 25 prompts that capture geometric shape (e.g., rectangular, circular), material appearance (e.g., white roofs, dark water), and spatial context (e.g., isolated placement, proximity to buildings). For example, we use ``\textit{a long white rectangular building located in an open, unobstructed area}'' for barn detection (see the Appendix Table~\ref{tab:infra_prompt_rationale_multirow}). To ensure that prompts are grounded in semantically valid regions and to reduce spurious detections, we apply infrastructure detection only to patches where metadata confirms the presence of the corresponding structure. Note that the prompts are not intended to infer the facility's animal type; rather, they target visual infrastructure elements that may appear across multiple CAFO categories. The 
    prompts are all listed in Table~\ref{tab:infra_prompt_rationale_multirow}.
    \item \textbf{Inpainting for Structure Removal}: For structure removal, we generate binary masks from subsets of predicted bounding boxes (up to five per image) and use these to guide the removal of detected infrastructure. The RGB image and corresponding mask are passed to a Stable Diffusion Inpainting model, conditioned on non-infrastructure prompts (e.g., “grassland”, “a small water pool”, “cluster of trees”). The model synthesizes contextually consistent content to fill the masked regions. To preserve the validity of the CAFO type label (e.g., swine, dairy, poultry), we remove infrastructure only when multiple instances are present within a patch. For example, if ten barns are detected, a subset (e.g., two or three) are removed to ensure the structural identity of the facility remains intact. This constraint prevents label ambiguity and preserves the semantic consistency of the augmented data. We also provide detailed metadata (including the number and type of structures removed, the inpainting prompt used, and the bounding box configuration) for each sample. This ensures full traceability and allows verification that label integrity is maintained across all synthetically generated samples.
\end{itemize}

This augmentation strategy introduces visual variation within the same semantic label, enabling models to generalize beyond repeated infrastructure patterns. By selectively removing structures while preserving the CAFO type, the model is encouraged to learn broader contextual features (e.g., landscape, vegetation, layout) rather than overfitting to infrastructure-specific cues.

\begin{table*}[t]
\caption{Grouped prompts and rationales for agricultural infrastructure types, with rationales structured by geometric shape, material appearance, and spatial context.}
\label{tab:infra_prompt_rationale_multirow}
\scriptsize
\centering
\renewcommand{\arraystretch}{1.3}
\setlength{\tabcolsep}{5pt}
\begin{tabular}{|p{1.5cm}|p{6cm}|p{6cm}|}
\hline
\textbf{Structure} & \textbf{Prompt(s)} & \textbf{Rationale} \\
\hline

\multirow{5}{*}{Barn} 
& a long white rectangular building located in an open, unobstructed area & 
\multirow{2}{6.2cm}{Rectangular in shape with elongated dimensions, featuring light-colored roofs, and positioned in isolated, open fields with minimal surrounding structures.} \\
\cline{2-2}
& a large rectangular structure placed alone in a spacious open area & \\
\cline{2-3}

& a light gray rectangular structure with a bright roof placed in a cleared field & 
\multirow{2}{6.2cm}{Rectangular buildings with reflective or light-colored roofs, typically situated in uniform, cleared agricultural areas.} \\
\cline{2-2}
& a rectangular building with a reflective or light-colored roof in a uniform open setting & \\
\cline{2-3}

& a simple white rectangular building with a flat or sloped roof surrounded by open land & 
\multirow{2}{6.2cm}{Wide, low-profile rectangular shapes with simple white roofs, located far from roads or residential structures.} \\
\cline{2-2}
& a wide, low-rise building with a simple roof and no nearby roads or houses & \\
\cline{2-3}

& a pale rectangular building that is isolated from other nearby structures & 
\multirow{2}{6.2cm}{Box-shaped buildings with pale or white surfaces, spatially detached from other infrastructure in cleared zones.} \\
\cline{2-2}
& a white box-shaped building standing by itself in a cleared landscape & \\
\cline{2-3}

& a long rectangular building with a white or gray roof and no surrounding clutter & 
\multirow{2}{6.2cm}{Linear rectangular structures with white or gray roofing, placed in bare or uncluttered terrain.} \\
\cline{2-2}
& a linear structure with strong roof edges, positioned in an empty ground space & \\
\hline

\multirow{4}{*}{Manure Pond} 
& a dark brown or black irregular pond surrounded by open land & 
\multirow{2}{6.2cm}{Irregularly shaped, dark-colored water bodies commonly located in bare land with limited vegetation or structures.} \\
\cline{2-2}
& a muddy or black spot on the ground surrounded by bare land & \\
\cline{2-3}

& a small black water pool next to a large building & 
\multirow{2}{6.2cm}{Dark, compact pools with irregular or circular shapes, typically near barns or placed away from residential zones.} \\
\cline{2-2}
& a dark patch of water placed away from houses or roads & \\
\cline{2-3}

& a dark green or brown pool of water near agricultural buildings & 
\multirow{2}{6.2cm}{Rectangular or rounded shapes with dark tones and low reflectance, often adjacent to farm structures.} \\
\cline{2-2}
& a rectangular dark spot with no reflection near farm infrastructure & \\
\cline{2-3}

& a dirty water pond placed in a fenced or open farm area & 
\multirow{2}{6.2cm}{Shallow, flat pits with dark water content, commonly found in fenced or exposed areas within farm boundaries.} \\
\cline{2-2}
& a small flat black or brown pit in open space & \\
\hline

\multirow{4}{*}{Other} 
& a tall round silver structure standing alone in an open area & 
\multirow{3}{6.2cm}{Cylindrical or rounded forms with metallic or reflective surfaces, typically isolated or aligned along roads or access routes.} \\
\cline{2-2}
& a silver or white round tank placed horizontally on the ground & \\
\cline{2-2}
& a dark or metallic cylinder standing upright near a road or dirt track & \\
\cline{2-3}

& a small box-shaped white structure placed near larger buildings & 
\multirow{2}{6.2cm}{Small rectangular or vertical pipe-like forms, white or neutral in appearance, located near barns or within farm complexes.} \\
\cline{2-2}
& a narrow vertical pipe-like structure standing on open soil & \\
\cline{2-3}

& a flat structure with no roof, made of metal or concrete, placed near a building & 
\multirow{2}{6.2cm}{Flat or elongated infrastructure lacking roofing, with metal or white coloring, typically positioned next to buildings.} \\
\cline{2-2}
& a narrow white object placed across open soil with no walls or roof & \\
\cline{2-3}

& a small building with a curved roof sitting in a field & 
\multirow{2}{6.2cm}{Curved or residential-style structures with white exteriors and driveways, indicative of greenhouses or farmhouses.} \\
\cline{2-2}
& a white rectangular building with windows and a driveway nearby & \\
\hline

\end{tabular}
\end{table*}

\section{Additional Details for Section~\ref{sec:benchmarking}}
\paragraph{Training Setup.} We trained all models using PyTorch on an NVIDIA A40 GPU with 40GB of memory. Training was conducted for a maximum of 30 epochs with early stopping (patience = 5) based on validation loss. We used a batch size of 32, a learning rate of 0.0001, and mixed-precision training (FP16) for improved efficiency. Each experiment used 12 data loading workers and was distributed across 2 GPUs where applicable. All models were trained on 8-class classification using the curated dataset described in Section~\ref{sec:benchmarking}.

\paragraph{Baseline Models}
In our benchmarking framework, we incorporate a diverse set of deep learning models to evaluate their effectiveness in both binary and multiclass CAFO classification tasks. We use ResNet18 and ResNet50 as strong convolutional baselines due to their proven reliability and computational efficiency. To explore transformer-based models, we include ViT-B/16 and Swin-B — with ViT offering global context understanding and Swin providing better spatial inductive biases through hierarchical attention. We also add EfficientNet-B0 and B3 for their state-of-the-art accuracy-efficiency tradeoffs, making them suitable for scalable deployment. Beyond these supervised architectures, we evaluate DINOv2-ViT-B, a self-supervised vision transformer known for its strong generalization on downstream tasks. Finally, we integrate CLIP and RemoteCLIP, two vision-language models that offer zero-shot capability and strong representation learning without task-specific fine-tuning.

\begin{figure*}[!ht]
    \centering
    \includegraphics[width=\textwidth]{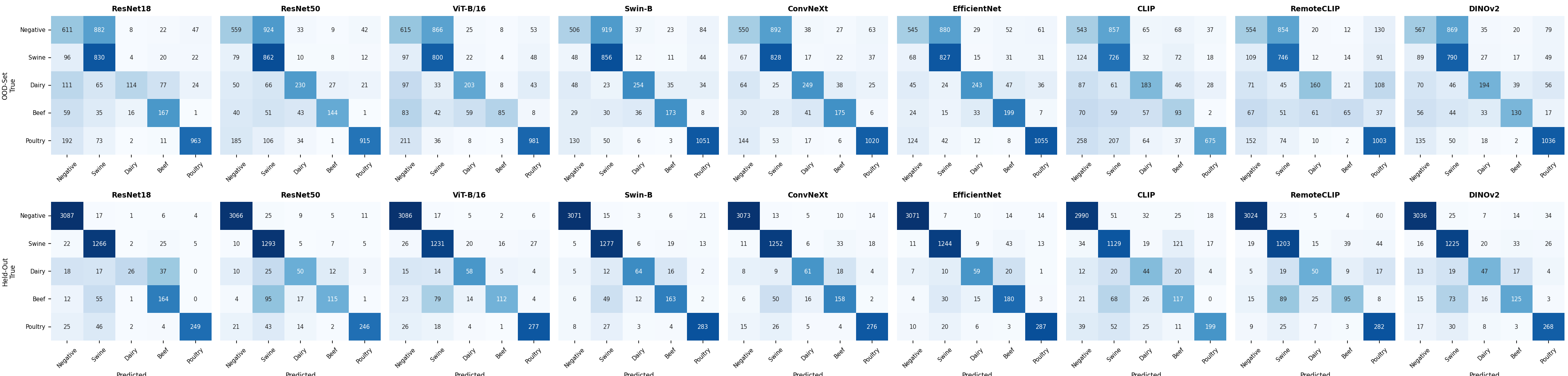}
    \caption{Confusion matrices for all nine models on the \texttt{Verified}-Set 
    (\textit{top row}) and Held-Out Set (\textit{bottom row}), restricted 
    to the five major classes. Cell values show raw counts; color intensity 
    reflects row-normalized accuracy. Models perform substantially better 
    on the in-distribution Held-Out Set. On the \texttt{Verified}-Set, Negative samples 
    are frequently misclassified as Swine, and \textit{Dairy} and 
    \textit{Beef} show persistent mutual confusion due to visual similarity. 
    \textit{Poultry} is the easiest class to recognize across both sets, 
    owing to its visually distinct long-barn infrastructure.}
    \label{fig:conf}
\end{figure*}

\begin{table*}[t]
\centering
\begin{tabular}{lccccc}
\toprule
\textbf{Model} & \textbf{Negative} & \textbf{Barn} & \textbf{Manure Pond} & \textbf{Grazing Area} & \textbf{Other Infra.} \\
\midrule
CLIP              & 0.432 & 0.505 & 0.414 & 0.313 & 0.689 \\
ConvNeXt-Tiny     & 0.406 & 0.589 & 0.494 & 0.563 & 0.825 \\
DINOv2-ViT-B      & 0.461 & 0.573 & 0.511 & 0.542 & 0.808 \\
EfficientNet-B0   & 0.463 & 0.593 & 0.618 & 0.593 & 0.863 \\
RemoteCLIP        & 0.465 & 0.515 & 0.385 & 0.343 & 0.717 \\
ResNet18          & 0.473 & 0.584 & 0.558 & 0.536 & 0.831 \\
ResNet50          & 0.464 & 0.611 & 0.579 & 0.574 & 0.842 \\
Swin-B            & 0.496 & 0.596 & 0.571 & 0.563 & 0.854 \\
ViT-B/16          & 0.483 & 0.591 & 0.593 & 0.546 & 0.836 \\
\bottomrule
\end{tabular}
\caption{Per-class F1 scores for infrastructure-level detection across all 
evaluated models, computed on the Verified Set. Each model is trained on 
the Standard Training Set and evaluated on binary presence/absence of 
barn, manure pond, grazing area, and other infrastructure within 
manually annotated CAFO patches. Bold values indicate the best 
per-column performance.}
\label{tab:cafo_classwise_f1}
\end{table*}

\paragraph{Evaluation Metrics}
To assess model performance, we use three complementary metrics: Accuracy, F1 Macro, and mean Average Precision (mAP). \textbf{Accuracy} measures the overall proportion of correct predictions across all classes but may be biased toward dominant classes in imbalanced datasets. To address this, we report \textbf{F1 Macro}, which computes the F1 score independently for each class and averages them, giving equal weight to all classes regardless of frequency—making it especially important for evaluating performance on underrepresented CAFO types. Finally, we include \textbf{mean Average Precision (mAP)} to capture the precision-recall trade-off across thresholds, providing a robust indicator of class separability and detection quality. Together, these metrics offer a comprehensive view of model performance across both dominant and minority classes.

%% file: main.bib
@String(CVPR= {IEEE Conf. Comput. Vis. Pattern Recog.})

@String(ICCV= {Int. Conf. Comput. Vis.})

@String(ICLR = {Int. Conf. Learn. Represent.})

@String(CVPR  = {CVPR})

@String(ICCV  = {ICCV})

@String(ICLR  = {ICLR})

@article{moses2017industrial,
  title={Industrial animal agriculture in the United States: Concentrated animal feeding operations (CAFOs)},
  author={Moses, Aurora and Tomaselli, Paige},
  journal={International farm animal, wildlife and food safety law},
  pages={185--214},
  year={2017},
  publisher={Springer}
}

@techreport{ehrenpreis2021nmcafo,
  author      = {Vanessa Ehrenpreis and Marshall Worsham and Nick Clarke and Adam Buchholz},
  title       = {Using Machine Learning to Map Concentrated Animal Feeding Operations in New Mexico},
  institution = {Mapping for Environmental Justice},
  year        = {2021},
  month       = {January},
  note        = {Report prepared for the McGovern Foundation},
  url         = {https://mappingforej.studentorg.berkeley.edu/wp-content/uploads/2022/03/NM-CAFO-Report.pdf}
}

@misc{nlcd,
  author = {Multi-Resolution Land Characteristics Consortium (MRLC)},
  title = {National Land Cover Database (NLCD)},
  year = {2021},
  url = {https://www.mrlc.gov/},
  note = {Accessed: 2024-03-01}
}

@article{handan2019deep,
  title={Deep learning to map concentrated animal feeding operations},
  author={Handan-Nader, Cassandra and Ho, Daniel E},
  journal={Nature Sustainability},
  volume={2},
  number={4},
  pages={298--306},
  year={2019},
  publisher={Nature Publishing Group UK London}
}

@article{zhu2022meter,
  title={METER-ML: a multi-sensor earth observation benchmark for automated methane source mapping},
  author={Zhu, Bryan and Lui, Nicholas and Irvin, Jeremy and Le, Jimmy and Tadwalkar, Sahil and Wang, Chenghao and Ouyang, Zutao and Liu, Frankie Y and Ng, Andrew Y and Jackson, Robert B},
  journal={arXiv preprint arXiv:2207.11166},
  year={2022}
}

@article{nguyen2025emergence,
  title={Emergence and interstate spread of highly pathogenic avian influenza A (H5N1) in dairy cattle in the United States},
  author={Nguyen, Thao-Quyen and Hutter, Carl R and Markin, Alexey and Thomas, Megan and Lantz, Kristina and Killian, Mary Lea and Janzen, Garrett M and Vijendran, Sriram and Wagle, Sanket and Inderski, Blake and others},
  journal={Science},
  volume={388},
  number={6745},
  pages={eadq0900},
  year={2025},
  publisher={American Association for the Advancement of Science}
}

@article{saha2025machine,
  title={Machine learning-based identification of animal feeding operations in the United States on a parcel-scale},
  author={Saha, Arghajeet and Rashid, Barira and Liu, Ting and Miralha, Lorrayne and Muenich, Rebecca L},
  journal={Science of The Total Environment},
  volume={960},
  pages={178312},
  year={2025},
  publisher={Elsevier}
}

@article{adiga2024high,
  title={A High-Resolution, US-scale Digital Similar of Interacting Livestock, Wild Birds, and Human Ecosystems with Applications to Multi-host Epidemic Spread},
  author={Adiga, Abhijin and Chopra, Ayush and Wilson, Mandy L and Ravi, SS and Xie, Dawen and Swarup, Samarth and Lewis, Bryan and Raskar, Ramesh and Marathe, Madhav V},
  journal={arXiv preprint arXiv:2411.01386},
  year={2024}
}

@article{prosser2024using,
    author = "Prosser, Diann J and Kent, Cody M and Sullivan, Jeffery D and Patyk, Kelly A and McCool, Mary-Jane and Torchetti, Mia Kim and Lantz, Kristina and Mullinax, Jennifer M",
    title = "Using an adaptive modeling framework to identify avian influenza spillover risk at the wild-domestic interface",
    journal = "Scientific Reports",
    volume = "14",
    number = "1",
    pages = "14199",
    year = "2024",
    publisher = "Nature Publishing Group UK London",
    original_key = "prosser2024using"
}

@article{humphreys2020waterfowl,
    author = "Humphreys, John M and Ramey, Andrew M and Douglas, David C and Mullinax, Jennifer M and Soos, Catherine and Link, Paul and Walther, Patrick and Prosser, Diann J",
    title = "Waterfowl occurrence and residence time as indicators of H5 and H7 avian influenza in North American Poultry",
    journal = "Scientific Reports",
    volume = "10",
    number = "1",
    pages = "2592",
    year = "2020",
    publisher = "Nature Publishing Group UK London",
    original_key = "humphreys2020waterfowl"
}

@ARTICLE{9832662,
  author={Robinson, Caleb and Chugg, Ben and Anderson, Brandon and Ferres, Juan M. Lavista and Ho, Daniel E.},
  journal={IEEE Journal of Selected Topics in Applied Earth Observations and Remote Sensing}, 
  title={Mapping Industrial Poultry Operations at Scale With Deep Learning and Aerial Imagery}, 
  year={2022},
  volume={15},
  number={},
  pages={7458-7471},
  doi={10.1109/JSTARS.2022.3191544}}

@article{CHUGG2021102463,
title = {Enhancing environmental enforcement with near real-time monitoring: Likelihood-based detection of structural expansion of intensive livestock farms},
journal = {International Journal of Applied Earth Observation and Geoinformation},
volume = {103},
pages = {102463},
year = {2021},
issn = {1569-8432},
doi = {https://doi.org/10.1016/j.jag.2021.102463},
url = {https://www.sciencedirect.com/science/article/pii/S0303243421001707},
author = {Ben Chugg and Brandon Anderson and Seiji Eicher and Sandy Lee and Daniel E. Ho}
}

@book{GurianSherman2008,
  author    = {D. Gurian-Sherman},
  title     = {{CAFOs Uncovered: The Untold Costs of Confined Animal Feeding Operations}},
  year      = {2008},
  publisher = {Union of Concerned Scientists}
}

@techreport{Hribar2010,
  author      = {C. Hribar},
  title       = {{Understanding Concentrated Animal Feeding Operations and Their Impact on Communities}},
  institution = {National Association of Local Boards of Health},
  year        = {2010}
}

@incollection{HandanNader2021,
  author    = {C. Handan-Nader and D. E. Ho and L. Y. Liu},
  title     = {{Deep Learning with Satellite Imagery to Enhance Environmental Enforcement}},
  booktitle = {{Data Science Applied to Sustainability Analysis}},
  pages     = {205--228},
  publisher = {Elsevier},
  year      = {2021}
}

@inproceedings{selvaraju2017grad,
  title={Grad-CAM: Visual explanations from deep networks via gradient-based localization},
  author={Selvaraju, Ramprasaath R. and Cogswell, Michael and Das, Abhishek and Vedantam, Ramakrishna and Parikh, Devi and Batra, Dhruv},
  booktitle={Proceedings of the IEEE International Conference on Computer Vision (ICCV)},
  pages={618--626},
  year={2017}
}

@article{li2022weakreview,
  title={Optical remote sensing image understanding with weak supervision: A comprehensive review},
  author={Li, Xiaoxiao and Tang, Zhihao and Xia, Gui-Song and Zhang, Liangpei},
  journal={arXiv preprint arXiv:2204.09120},
  year={2022}
}

@article{zhu2017deep,
  title={Deep learning in remote sensing: A comprehensive review and list of resources},
  author={Zhu, Xiaoxiang and Tuia, Devis and Mou, Lichao and Xia, Gui-Song and Zhang, Liangpei and Xu, Feng and Fraundorfer, Friedrich},
  journal={IEEE Geoscience and Remote Sensing Magazine},
  volume={5},
  number={4},
  pages={8--36},
  year={2017},
  publisher={IEEE}
}

@article{liu2023groundingdino,
  title={Grounding DINO: Marrying DINO with grounded pre-training for open-set object detection},
  author={Liu, Shilong and Li, Feng and Wu, Hao and others},
  journal={arXiv preprint arXiv:2303.05499},
  year={2023}
}

@misc{usda_naip,
  author       = {{U.S. Department of Agriculture}},
  title        = {{National Agriculture Imagery Program (NAIP)}},
  year         = {2023},
  note         = {\url{https://www.fsa.usda.gov/programs-and-services/aerial-photography/imagery-programs/naip-imagery/}},
  howpublished = {\url{https://www.fsa.usda.gov/programs-and-services/aerial-photography/imagery-programs/naip-imagery/}},
  institution  = {USDA Farm Service Agency},
}

@inproceedings{rombach2022high,
  title={High-resolution image synthesis with latent diffusion models},
  author={Rombach, Robin and Blattmann, Andreas and Lorenz, Dominik and Esser, Patrick and Ommer, Bj{\"o}rn},
  booktitle={Proceedings of the IEEE/CVF Conference on Computer Vision and Pattern Recognition},
  pages={10684--10695},
  year={2022}
}

@article{wang2023promptdiffusion,
  title={PromptDiffusion: Generating large-scale datasets with subject-prompt alignment},
  author={Wang, Yi and Tan, Hao and Yu, Hang},
  journal={arXiv preprint arXiv:2308.07974},
  year={2023}
}

@article{chen2023diffusionaug,
  title={Diffusion-Aug: Unlocking the Power of Diffusion Models for Visual Recognition},
  author={Chen, Ting and Li, Xinyang and Sohn, Kihyuk and Wang, Zizhao and Yuan, Lu and Zhang, Han},
  journal={arXiv preprint arXiv:2302.07685},
  year={2023}
}

@misc{cafomaps,
  author       = {{Department of Geographical and Sustainability Sciences, University of Iowa}},
  title        = {CAFOMaps: Concentrated Animal Feeding Operations in the United States},
  year         = {2025},
  howpublished = {\url{https://www.cafomaps.org/}},
  note         = {Accessed: 2025-05-16}
}

@misc{NYcafo,
  author       = {{New York Department of State}},
  title        = {New York CAFO Dataset},
  year         = {2024},
  howpublished = {\url{https://opdgig.dos.ny.gov/datasets/a9a8eaed80864ab98680899ecdbc1c50/explore?location=42.664852\%2C-76.586900\%2C7.22}},
  note         = {Accessed: 2024-05-15}
}

@misc{MNcafo,
  author       = {Minnesota Geospatial Commons},
  title        = {Minnesota CAFO Locations (Layer 0)},
  year         = {2024},
  howpublished = {\url{https://www.arcgis.com/home/item.html?id=6d119156229d4e908e22f027bdaee6be\&sublayer=0}},
  note         = {Accessed: 2024-05-15}
}

@misc{MIcafo,
  author       = {{Michigan EGLE GIS Hub}},
  title        = {CAFO Locations - Michigan Department of Environment, Great Lakes, and Energy},
  year         = {2024},
  howpublished = {\url{https://gis-egle.hub.arcgis.com/datasets/f0843875e5874d04b06396de8200cf75/explore?location=43.005451\%2C-83.963570\%2C6.10}},
  note         = {Accessed: 2024-05-15}
}

@misc{MDcafo,
  author       = {Maryland Department of the Environment},
  title        = {Animal Feeding Operations Map},
  year         = {2024},
  howpublished = {\url{https://catalog.data.gov/dataset/maryland-department-of-the-environment-lma-resource-management-program-animal-feeding-oper-5fb42}},
  note         = {Accessed: 2024-05-15}
}

@misc{IAcafo,
  author       = {Iowa Geodata Portal},
  title        = {Iowa Animal Feeding Operations GIS Data},
  year         = {2024},
  howpublished = {\url{https://geodata.iowa.gov/documents/abfbd972640d4e87b6c48dc669775767/about}},
  note         = {Accessed: 2024-05-15}
}

@misc{INcafo,
  author       = {IndianaMap},
  title        = {Confined Feeding Operations - IndianaMap},
  year         = {2024},
  howpublished = {\url{https://www.indianamap.org/datasets/INMap::confined-feeding-operations/about}},
  note         = {Accessed: 2024-05-15}
}

@misc{DEcafo,
  author       = {Delaware Department of Natural Resources},
  title        = {Delaware CAFO Map Viewer},
  year         = {2024},
  howpublished = {\url{https://experience.arcgis.com/experience/c6749f8b31d143cbb38a26fc1b89a2be/page/Delaware}},
  note         = {Accessed: 2024-05-15}
}

@article{he2016deep,
  title={Deep Residual Learning for Image Recognition},
  author={He, Kaiming and Zhang, Xiangyu and Ren, Shaoqing and Sun, Jian},
  journal={CVPR},
  year={2016}
}

@article{tan2019efficientnet,
  title={EfficientNet: Rethinking Model Scaling for Convolutional Neural Networks},
  author={Tan, Mingxing and Le, Quoc},
  journal={ICML},
  year={2019}
}

@article{liu2022convnet,
  title={A ConvNet for the 2020s},
  author={Liu, Zhuang and Mao, Hanzi and Wu, Chao-Yuan and Feichtenhofer, Christoph and Darrell, Trevor and Xie, Saining},
  journal={CVPR},
  year={2022}
}

@article{dosovitskiy2020vit,
  title={An Image is Worth 16x16 Words: Transformers for Image Recognition at Scale},
  author={Dosovitskiy, Alexey and Beyer, Lucas and Kolesnikov, Alexander and Weissenborn, Dirk and Zhai, Xiaohua and Unterthiner, Thomas and Dehghani, Mostafa and Minderer, Matthias and Heigold, Georg and Gelly, Sylvain and Uszkoreit, Jakob and Houlsby, Neil},
  journal={ICLR},
  year={2021}
}

@article{liu2021swin,
  title={Swin Transformer: Hierarchical Vision Transformer using Shifted Windows},
  author={Liu, Ze and Lin, Yutong and Cao, Yuqi and Hu, Han and Wei, Yixuan and Zhang, Zheng and Lin, Stephen and Guo, Baining},
  journal={ICCV},
  year={2021}
}

@article{oquab2023dinov2,
  title={DINOv2: Learning Robust Visual Features without Supervision},
  author={Oquab, Maxime and Darcet, Thomas and Moutakanni, Theo and Ram{\'e}, Alexandre and Haziza, Daniel and Su{\'a}rez, Jorge and Szafraniec, Marc and Kalantidis, Yannis and Elkabetz, Yair and Cord, Matthieu and others},
  journal={arXiv preprint arXiv:2304.07193},
  year={2023}
}

@article{radford2021clip,
  title={Learning Transferable Visual Models From Natural Language Supervision},
  author={Radford, Alec and Kim, Jong Wook and Hallacy, Christopher and Ramesh, Aditya and Goh, Gabriel and Agarwal, Sandhini and Sastry, Girish and Askell, Amanda and Mishkin, Pamela and Clark, Jack and others},
  journal={ICML},
  year={2021}
}

@inproceedings{mallya2022remoteclip,
  title={Learning from Noisy Remote Sensing Data with Vision-Language Models},
  author={Mallya, Arun and Chen, Guangxing and Xie, Weicheng and Zhu, Xiaojin and Lobell, David and Ermon, Stefano},
  booktitle={NeurIPS},
  year={2022}
}

@article{khanna2023diffusionsat,
  title={Diffusionsat: A generative foundation model for satellite imagery},
  author={Khanna, Samar and Liu, Patrick and Zhou, Linqi and Meng, Chenlin and Rombach, Robin and Burke, Marshall and Lobell, David and Ermon, Stefano},
  journal={arXiv preprint arXiv:2312.03606},
  year={2023}
}

@inproceedings{sastry2024geosynth,
  title={Geosynth: Contextually-aware high-resolution satellite image synthesis},
  author={Sastry, Srikumar and Khanal, Subash and Dhakal, Aayush and Jacobs, Nathan},
  booktitle={Proceedings of the IEEE/CVF Conference on Computer Vision and Pattern Recognition},
  pages={460--470},
  year={2024}
}

@article{de2024data,
  title={Data Augmentation in Earth Observation: A Diffusion Model Approach},
  author={DE JESUS SOUSA, Tiago Alexandre and RIES, Benoit and GUELFI, Nicolas},
  year={2024}
}

@InProceedings{rombach2022ldm,
    author    = {Rombach, Robin and Blattmann, Andreas and Lorenz, Dominik and Esser, Patrick and Ommer, Bj\"orn},
    title     = {High-Resolution Image Synthesis With Latent Diffusion Models},
    booktitle = {Proceedings of the IEEE/CVF Conference on Computer Vision and Pattern Recognition (CVPR)},
    month     = {June},
    year      = {2022},
    pages     = {10684-10695}
}

@article{nguyen2024generating,
  title={Generating Synthetic Satellite Imagery for Rare Objects: An Empirical Comparison of Models and Metrics},
  author={Nguyen, Tuong Vy and Hoster, Johannes and Glaser, Alexander and Hildebrand, Kristian and Biessmann, Felix},
  journal={arXiv preprint arXiv:2409.01138},
  year={2024}
}

@misc{fernandez2020torchcam,
  title     = {{TorchCAM}: Class Activation Explorer},
  author    = {Fran{\c{c}}ois-Guillaume Fernandez},
  year      = {2020},
  month     = {March},
  publisher = {GitHub},
  howpublished = {\url{https://github.com/frgfm/torch-cam}}
}
